\documentclass{article}
\usepackage{spconf,amsmath,graphicx}
\usepackage{subfigure}
\usepackage{booktabs}
\usepackage{amsfonts}
\usepackage{bm}
\usepackage{amsmath}

\usepackage{enumitem}
\setlist{nosep, leftmargin=14pt}

\usepackage{mwe} 
\usepackage{cases}

\title{Decoding Decision Reasoning: A Counterfactual-Powered Model for Knowledge Discovery}

\name{Yingying Fang$^{1}$, Zihao Jin$^{2}$, Xiaodan Xing$^{3}$, Simon Walsh$^{1}$, Guang Yang$^{1,3}$}
\address{$^{1}$ \small{National Heart and Lung Institute, Imperial College London, London, UK} \\
$^{2}$ \small{Department of Metabolism, Digestion and Reproduction, Imperial College London, London, UK}\\
$^{3}$ \small{Bioengineering Department and Imperial-X, Imperial College London, London, UK}}

\begin{document}

\maketitle

\begin{abstract}

In medical imaging, particularly in early disease detection and prognosis tasks, discerning the rationale behind an AI model's predictions is crucial for evaluating the reliability of its decisions.  Conventional explanation methods  face challenges in identifying discernible decisive features in medical image classifications, where discriminative features are subtle or not immediately apparent.
To bridge this gap, we propose an explainable model that is equipped with both decision reasoning and feature identification capabilities. Our approach not only detects influential image patterns but also uncovers the decisive features that drive the model's final predictions. By implementing our method, we can efficiently identify and visualise class-specific features leveraged by the data-driven model, providing insights into the decision-making processes of deep learning models. We  validated our model in the demanding realm of medical prognosis task, demonstrating its efficacy and potential in enhancing the reliability of AI in healthcare and in discovering new knowledge in diseases where prognostic understanding is limited.
\end{abstract}

\section{Introduction}
\label{sec:intro}

The integration of artificial intelligence (AI) into medical imaging has significantly transformed patient care, offering a cost-effective and widely accessible means of enhancing diagnostic and prognostic accuracy. However, the  `black-box' nature of decision-making processes in current AI-based medical imaging models hinders their practical application in real-world scenarios. Addressing this challenge, explainable AI (XAI) techniques have been developed over the last few decades to illuminate the inner workings of these models.

Traditional XAI methods, such as backpropagation-based methods \cite{sundararajan2017axiomatic} and activation-based methods \cite{selvaraju2017grad, li2022explainable}, typically create attribution maps to highlight areas significant to a model's decisions.
However, in the context of high-stakes medical image classification, pinpointing specific attributes for each decision is crucial. Attribution maps, focusing only on decision-related regions, struggle to identify distinct features for each classification in scenarios where key areas overlap significantly across classes \cite{rudin2019stop}.
 
In contrast, a counterfactual explanation proposed by Wachter \textit{et al.} \cite{wachter2017counterfactual} provides a distinctive approach to explaining the models by generating contrastive examples that can alter the model's decision based on identified features. The comparison between the original image and its counterfactual counterpart facilitates the identification of critical features influencing the model's predictions. 
The recent evolution of generative AI models, particularly GAN-based approaches, has led to their successful application in providing counterfactual explanations for medical images \cite{atad2022chexplaining, mertes2022ganterfactual, singla2023explaining, schutte2021using,sankaranarayanan2022real}.  These methods stand out in producing highly realistic counterfactual examples  with alterations that are easily identifiable by humans. 
Compared to the earlier XAI techniques, these methods provide more informative  and clear explanations by directly modifying the influential features in the image.
However, despite their success in generating counterfactual images, these methods are designed to explain a pre-trained `black box' model rather than being applied to provide an intrinsically interpretable model. Besides, these methods entail significant computational costs in pinpointing the most influential features, as they rely on generating a large quantity of candidate counterfactuals and subsequently ranking the influence of all potential changes.

Differently, we present a novel intrinsically explainable prognosis model which is powered with counterfactual generations, named PrognosisEx. 
Specifically,  we enhance the explainability of the classification process by decoding the reasoning behind its decisions and visualising the `decisive' features.
The key contributions of this study include: 
(1) we introduce a diffusion-based autoencoder for chest CT slices, capable of generating semantically meaningful representations and reconstructing the original slices with these representations; 
(2) we propose a reasoning space for decoding of the model's decision and understanding the contributing features within the developed interpretable model; 
(3) we propose a computationally efficient counterfactual generation method for semantic understanding of identified features in images, which will identify the crucial features before generating the counterfactuals,
(4) we demonstrate the efficacy of PrognosisEx in biomarker discovery when applied to predicting the 10-day mortality of COVID-19 inpatients.

\section{Methodology}
\label{sec:Method}

\begin{figure}[htb]
\centerline{\includegraphics[trim=0 0 0 0,clip,width=0.8\linewidth]{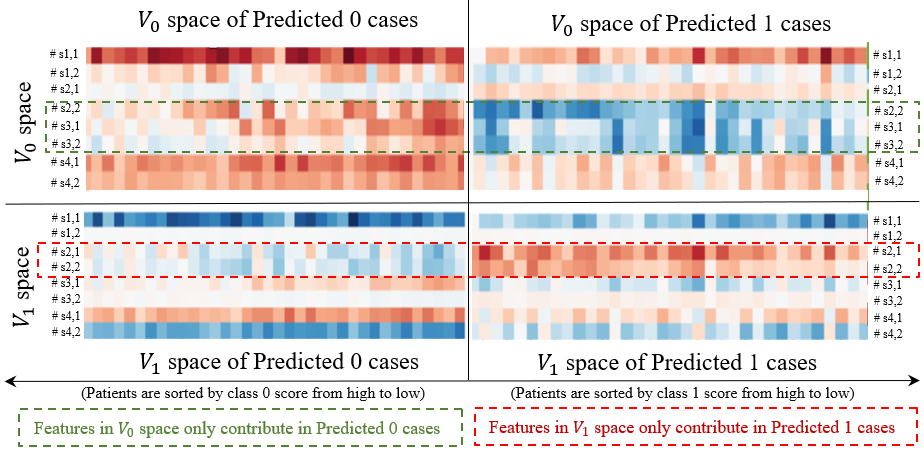}}
\caption{Identification of class-specific attributes in $\mathbb{V}_c$ space for our binary prognostic task to predict the mortality of COVID-19 patients in 10 days, where class 1 represents the label of death and class 0 represents the label of survival. In this case, $s_{2,1}$ and $s_{2,2}$ in $\mathbb{V}_1$ space  are class 1-specific features, which are regarded as most decisive features to decision 1. Similarly, $s_{2,2}$, $s_{3,1}$, and $s_{3,2}$ in $\mathbb{V}_0$ space are class 0-specific features, considered the most decisive features to decision 0.}
\label{decoding}
\end{figure}

The complete workflow of PrognosisEx for decoding the model's decisions is depicted in Fig.~\ref{Framework} (A), and the pretraining process of each module is presented in Fig.~\ref{Framework} (B).
The proposed PrognosisEx decodes the model's decisions by first identifying the decisive features within the model, and then identifying their semantic features from images.
PrognosisEx begins by using the encoder from the pretrained autoencoder to extract the representations from CT slices, followed by a two-layer prognostic model. 
The first layer of the prognostic model generates features from  representations of each slice. This feature space allows us to pinpoint the most influential features for the model's prediction for each class, as depicted in Fig.~\ref{decoding}.
As the final step, we manipulate these key features within their respective representations and create
`counterfactual examples',  either by amplifying or diminishing their attributes.
By contrasting these counterfactual images with the original slices, we  discern the semantic meaning of 
the identified features.

\subsection{Diffusion-based autoencoder}
\label{diffusion}

\begin{figure}[t]
\centerline{\includegraphics[width=1\linewidth]{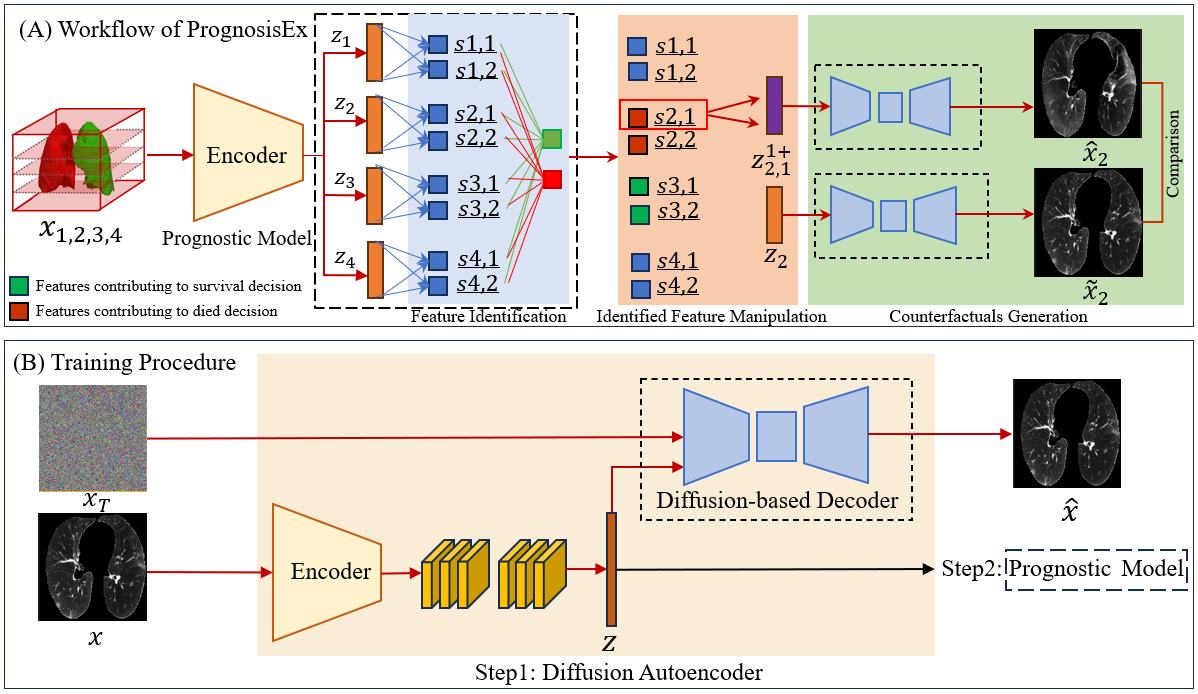}}  
\caption{(A) The decision-decoding workflow of PrognosisEx; (B) The training process of PrognosisEx, which includes the development of a general autoencoder that compresses slices into a low-dimensional vector, followed by the training of a two-layer classifier model.
}
\label{Framework}
\end{figure}

As depicted in Fig.~\ref{Framework} (B), 
the representation in our autoencoder serves two crucial roles in achieving our interpretable prognosis model. First, it functions as a foundation for downstream prognosis tasks. Second, it forms the basis for controlling the content of generated images, serving as the manipulated space for generating counterfactual explanations. For these purposes, we find the DDIM autoencoder introduced  in \cite{preechakul2022diffusion} to be an ideal option, which surpasses other generative models in terms of both image reconstruction and image generation abilities, while also possessing a semantically meaningful representation space.

Our autoencoder model, taking an individual slice as its input data, consists of two main parts: an encoder $\mathbf{E}_{\phi}$ and a decoder $\mathbf{D}_{\theta}$:
the encoder compresses the high-dimensional image $\mathbf{x}_0$ into a semantically meaningful low-dimensional representation $\mathbf{z} \in \mathbb{R}^{512}$, while the decoder generates the reconstructed slice $\hat{\mathbf{x}}_0$ from $\mathbf{z}$. The reconstruction is achieved  by iteratively performing the following step with a conditional diffusion model $\mathbf{D}_{\theta}$:
\begin{multline}
    \hat{\mathbf{x}}_{t-1}=\sqrt{\alpha_{t-1}}\left(\frac{\hat{\mathbf{x}}_t-\sqrt{1-\alpha_t} \mathbf{D}_\theta\left(\hat{\mathbf{x}}_t, t, \mathbf{z}\right)}{\sqrt{\alpha_t}}\right)\\+\sqrt{1-\alpha_{t-1}} \mathbf{D}_\theta\left(\hat{\mathbf{x}}_t, t, \mathbf{z}\right), \text{with } t=T,T-1,...,1
\end{multline}
where $\hat{\mathbf{x}}_{T}= \mathbf{x}_{T} = \sqrt{\alpha_T} \boldsymbol{x}_0+\sqrt{1-\alpha_T} \boldsymbol{\epsilon}$ and $\boldsymbol{\epsilon} \sim \mathcal{N}(\mathbf{0}, \mathbf{I})$.
 
Following Preechakul \textit{et al.} \cite{preechakul2022diffusion}, the  models $\mathbf{E}$ and $\mathbf{D}$ are trained in tandem through optimizing the following loss function with respect to $\theta$ and $\phi$:
\begin{equation}
\mathcal{L}_{(\phi, \theta)}=\mathbb{E}\left\|\boldsymbol{\epsilon}-\mathbf{D}_\theta\left(\mathbf{x}_t, t, \mathbf{E}_\phi(\mathbf{x}_0)\right)\right\|_1.
\end{equation}

\subsection{Two-layer prognostic model}
In lung diseases such as COVID-19 and fibrotic lung disease, abnormalities usually manifest at varying lung locations. To circumvent manual slice selection for each patient and prevent overfitting from whole-scan input, our approach randomly selects $N$ slices from $N$ equivalently spaced areas from a complete lung scan per patient.  PrognosisEx's architecture, depicted in Fig.~\ref{Framework} (A), integrates the $N$ representations from the encoder  $\{\mathbf{z_n}\}$ as the input of the two-layer linear model.

The first layer comprises $N$ individual linear models $\mathbf{W}^{(1,n)}\in \mathbb{R}^{M,512}$, which are designed to extract $M$ features per slice, denoted as ${s}_{n,m}$. The second decision layer, with $\mathbf{W}^{(2,c)} \in \mathbb{R}^{1,MN}$ for each class $c$, aggregates these features across all slices into a single feature vector $\mathbf{s} = [{s}_{1,1}, ..., {s}_{1,M}, ... , s_{N,M}]$, and generates a final score $y_c$. The prognostic model is trained with a weighted objective function to improve classification of challenging samples:
\begin{equation}
\mathcal{L}_{\text{cls}}= -(1 - p_i)\log(p_i) 
\text{ with } p_i = \frac{e^{y_i}}{\sum_{j=1}^{C} e^{y_j}},
\end{equation}
where $i$ is the index of the true class for the given sample.

\subsection{Explainability of PrognosisEx}
The model's explainability is realised in two distinct ways: (1) its capacity to reason its decisions and identify the decisive features within the model; and (2) finding the semantic meaning of identified features  by generating counterfactual examples through a diffusion-based decoder,  both of which are integral to decoding decision reasoning..

\noindent{\textbf{Decision reasoning.}} To reason about the  model's decisions, we propose a reasoning space $\mathbb{V}_c$, consisting of  decision vectors that transformed from the extracted feature and decision layer $\mathbf{v}^{(c)} = \mathbf{W}^{(2,c)}\text{Diag}(\mathbf{s}) = [W^{(2,c)}_{1}{s}_{1,1},.., W^{(2,c)}_{NM}{s}_{N,M}]^{T}$, where ${v}^{(c)}_{n,m} = W^{(2,c)}_{nm}s_{n,m}$ is the contribution of $s_{n,m}$ in the prediction score of class $c$.
The intent behind $\mathbb{V}_c$ is to distil the model's complex reasoning into simpler, class-specific vectors that underscore the distinctive features between different classes. In the proposed space $\mathbb{V}_c$ for class $c$, we can detect the features that exclusively appear in cases predicted as class $c$. 
This allows us to attribute the model's prediction of class $c$ to the presence of these features in a predicted case, which are referred to as  `class-specific' features throughout this paper.
An example of decision reasoning approach is demonstrated in Fig.~\ref{decoding}, where we apply our method to a binary PrognosisEx with four slices and two features. 
From its visualisation, we can clearly observe that: (a) the features in $\mathbb{V}_c$ demonstrate remarkable consistent distribution within the same class across different individuals and (b) features in $\mathbb{V}_c$  provide clear differentiation between different classes.
These observations support the reasonable use of `class-specific' features for reasoning the model's decision.

\noindent\textbf{Counterfactual generation.} The final goal of PrognosisEx is to identify the semantic meaning of these pinpointed `class-specific' features though their visualisation on the original images. To achieve this, we adopt counterfactual explanation methods to discern crucial features. 
Specifically, we strive to create counterfactual slices 
from representations with either amplified or reduced contributions of `class-specific' features, thereby revealing the impact of identified features. Since that counterfactual image  are synthesised by the decoder from the $\mathbf{z}$ space from the trained autoencoder, 
we propose a method for manipulating ${\mathbf{z}_n}$, which can result in achieving our manipulation objectives of enhancing or mitigating the contribution of the target $s{n,m}$ to the prediction score for $c$, as follows
\begin{numcases}\\
\mathbf{z}^{c+}_{n,m} = \mathbf{z_n} + \alpha\cdot\text{sgn}(W^{(2,c)}_{nm})\cdot\nabla_{\mathbf{z}_n} s_{n,m} \label{enhance}\\ 
\mathbf{z}^{c-}_{n,m} = \mathbf{z_n} - \alpha\cdot\text{sgn}(W^{(2,c)}_{nm})\cdot\nabla_{\mathbf{z}_n} s_{n,m}. \label{decrease}
\end{numcases}
where  $\alpha>0$ is the stepsize. The corresponding enhanced and mitigated contribution of $s_{n,m}$ in prediction for class $c$ and  generated counterfactuals can be derived as 
$ \Tilde{v}^{(c)}_{n,m}={v}^{(c)}_{n,m}\pm\alpha|W_{n,m}^{(2,c)}|\|\mathbf{W}^{(1,n)}_m\|_2$ and $\Tilde{x}_{n} = \mathbf{D}_\theta(\mathbf{z}^{c\pm}_{n,m}).$

\section{Experimental Results}
\label{sec:Experiments}

\begin{figure*}[t] 
\centering 
\includegraphics[width=0.95\linewidth]{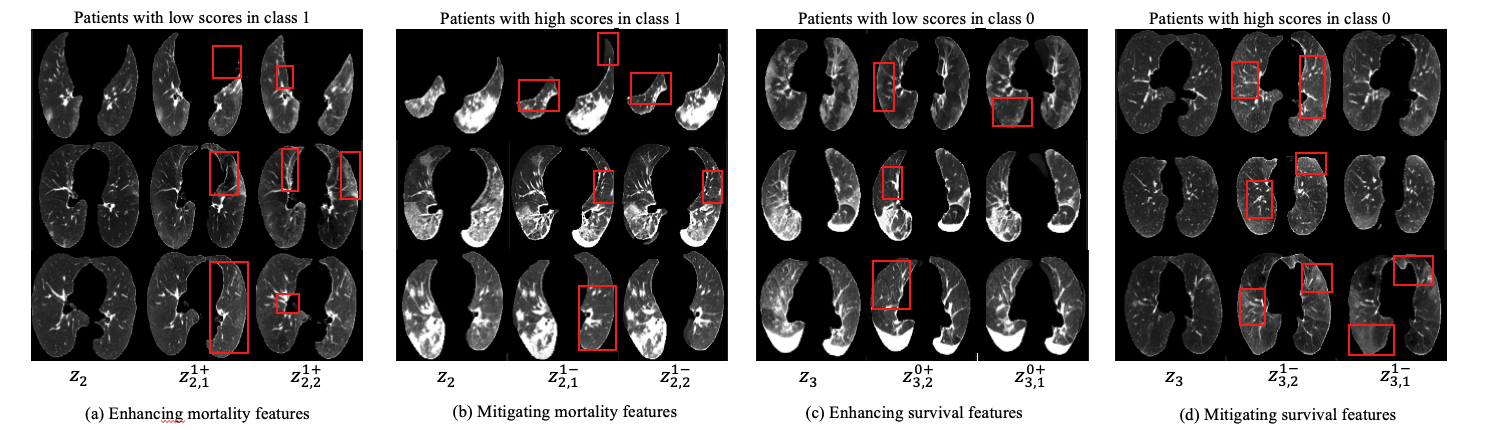}
\caption{Visualisation of identified `class-specific' features in the COVID-19 prognostic task using PrognosisEx. Figure (a) and (b) depict counterfactuals demonstrating the  enhanced and mitigated contribution of `Class 1-Specific' Features (features contributing to the death decision); Figure (c) and (d) depict counterfactuals demonstrating the enhanced and mitigated influence of `Class 0-Specific' features (features contributing to survival decision). In each image, the 1st column shows the image reconstructed from the unmanipulated feature, with the following two columns the images reconstructed from the manipulated features.}
\label{results}
\end{figure*}

Since there is no ground truth for explanation methods, our experiments are primarily aimed at demonstrating whether the model effectively decodes its decisions and identifies significant image patterns related to those decisions.
To further highlight the superiority of our proposed PrognosisEx model, we also conduct comparative analyses with other leading end-to-end deep learning models \cite{he2016deep,tan2019efficientnet,yu2022metaformer,huang2017densely}.

\subsection{Implementation Details}

\noindent\textbf{Dataset.} 
We validated the performance of PrognosisEx on our in-house dataset for a binary prognostic task to predict the 10-day mortality of COVID-19 inpatients. This dataset consists of 566 COVID-19 inpatients from the University Hospital of Parma, including 257 patients who died within 10 days and 309 patients who survived beyond 10 days during their hospitalisation.

\noindent\textbf{Data processing.} 
We divided the  complete lung equally into 4 regions from top to bottom and randomly selected four slices as the input of all models. 
To train and test our model, we employ patient-level data splitting, reserving 20\% for testing and using the rest for five-fold cross-validation. In each training, we trained the model from scratch with four folds used for training and one fold for validating.  
The area under the curve (AUC), accuracy, sensitivity, specificity, and weighted Youden’s index $J_{w}$ \cite{li2013weighted} were used as metrics for evaluating the prognostic performance. The one with the highest AUC on the validation set is selected as the final model for testing.\\
\noindent\textbf{Training details.} 
The autoencoder, with an input size of 256$\times$ 256, was trained on 45,302 slices using eight V100 GPUs with a learning rate of $1e^{-4}$ and a batch size of 64 for 100 epochs. PrognosisEx was set with $N=4$, $M=2$.  The end-to-end models and PrognosisEx were both trained using the Adam optimizer on an RTX3090 GPU with a batch size of 8 for 70 epochs with learning rates of 1e-6 and 1e-5, respectively, to achieve their best performance.

\subsection{Decoding decision reasoning}
In this section, we decode the model's decision by understanding the semantic meaning of the `class-specific' features identified in Fig.~\ref{decoding}. 
Specifically, we analyse the counterfactuals  from manipulating class 1 specific features  ($s_{2,1}$ and $s_{2,2}$), to understand the features leveraged by the model to increase the mortality probability, and counterfactuals  from  $s_{3,1}$ and $s_{3,2}$ to understand the semantic features contributing to higher probability of survival.
As Fig.~\ref{results}(a) demonstrates, manipulating 
$\mathbf{z}_{2,1}^{1+}$ and $\mathbf{z}_{2,2}^{1+}$
  to increase the death score by Eq.~\eqref{enhance}, leads to increased vascularisation, more ground-glass opacities (GGOs), and a reduced lung area in the second slice compared to the original slices. Consistently, reducing these features  
 $\mathbf{z}_{2,1}^{1-}$ and $\mathbf{z}_{2,2}^{1-}$ through Eq.~\eqref{decrease} presents the opposite effect, as shown in Fig.~\ref{results}(b). Using the same method, we unveil that the features contributing to higher probability of `survival' are associated with fewer vessel textures and higher opacities in third slice. Combining these feature semantics with the location of the manipulated slice, we conclude that features such as GGO, increased vessels, and diminished lung areas signal that appeared in the upper middle lung, increased mortality risk to the model. Conversely, fewer vessel textures and reduced opacity in third slices suggest higher survival odds. This demonstrates the knowledge that the model acquired to differentiate risk across pulmonary features.

\subsection{Comparison to End-to-end network}
The performance comparison of PrognosisEx with other leading models is presented in Table \ref{tab:performance}. PrognosisEx achieves comparable COVID-19 prognostic performance to deep black-box models, demonstrating its superior explainability without compromising state-of-the-art prognostic accuracy.

\begin{table}[h]
\centering
\caption{Performance of deep learning models on the test set.}
\label{tab:performance}
\renewcommand\arraystretch{1.1}
\resizebox{1\linewidth}{!}{
\begin{tabular}{c|cccccc}
\toprule[1.5pt]
Method & AUC & Accuracy & Sensitivity & Specificity & $J_{0.5}$ & $J_{0.6}$ \\ \hline
Resnet18 \cite{he2016deep}  & 0.77 & 0.71 & 0.68  & 0.74 & 0.71 & 0.70 \\ \hline
Resnet50 \cite{he2016deep}  & 0.71 & 0.69 & 0.75  & 0.63 & 0.69 & 0.70 \\ \hline
$\text{EfficientNet\_b0}$ \cite{tan2019efficientnet}  & 0.71 & 0.62 & 0.58  & 0.65 & 0.62 & 0.61 \\ \hline
$\text{PoolFormer\_v2\_tiny}$ \cite{yu2022metaformer}  & 0.69 & 0.62 & 0.62  & 0.63 & 0.62 & 0.62 \\ \hline
Densenet-121 \cite{huang2017densely} & 0.79 & 0.71 & 0.65  & 0.76 & 0.71 & 0.70 \\ \hline
PrognosisEx & 0.79 & 0.71 & 0.71  & 0.71 & 0.71 & 0.71\\
\bottomrule[2pt]
\end{tabular}
}
\end{table}

\section{Discussion and Conclusion}
\label{sec:Conclusion}
We further investigated whether the features identified by PrognosisEx align with existing prognostic knowledge of the disease. Notably, studies [21, 22, 23] indicated that the features used by the model for decision-making strongly correlate with biomarkers recognised in their research, suggesting that PrognosisEx could serve as an effective and cost-efficient approach to biomarker discovery, particularly in areas where prognostic knowledge is scarce.

In conclusion, this work presents an innovative, explainable prognostic model, which offers insight into its decision process and identifying features utilised for decision. Explainability in these two aspects notably improve the safety of applying AI models in healthcare. Besides, PrognosisEx demonstrates promise as an inexpensive tool for biomarker discovery and bias detection, especially in contexts lacking extensive prior knowledge.
In future work we will provide a broader validation  of the proposed method with different imaging modalities across various medical scenarios including the public datasets.

\section{COMPLIANCE WITH ETHICAL STANDARDS}
The dataset used in this study  was obtained from the University Hospital of Parma, following the approval of the local Ethics Committee (code 934/2021/OSS/AOUPR - 11.01.2022) is from the University Hospital of Parma.

\section{ACKNOWLEDGMENT}
This study was supported in part by the ERC IMI (101005122), the H2020 (952172), the MRC (MC/PC/21013), the Royal Society (IEC/NSFC/211235), the NVIDIA Academic Hardware Grant Program, the SABER project supported by Boehringer Ingelheim Ltd, Wellcome Leap Dynamic Resilience, NIHR Imperial Biomedical Research Centre, and the UKRI Future Leaders Fellowship(MR/V023799/1).

\bibliographystyle{IEEEbib}
\bibliography{strings,refs}
\end{document}